\documentclass[10pt,twocolumn,letterpaper]{article}
\usepackage[pagenumbers]{cvpr} 

\usepackage{graphicx}
\usepackage{amsmath}
\usepackage{amssymb}
\usepackage{booktabs}

\usepackage{multirow} 
\usepackage{xspace} 
\usepackage{soul} 
\usepackage[percent]{overpic} 

\makeatletter
\@namedef{ver@everyshi.sty}{}
\makeatother
\usepackage{tikz}

\usepackage[utf8]{inputenc}
\usepackage{pgfplots}
\DeclareUnicodeCharacter{2212}{−}
\usepgfplotslibrary{groupplots,dateplot}
\usetikzlibrary{patterns,shapes.arrows}
\pgfplotsset{compat=newest}

\DeclareMathOperator*{\argmin}{argmin}
\DeclareMathOperator*{\argmax}{argmax}

\usepackage[pagebackref,breaklinks,colorlinks]{hyperref}

\usepackage[capitalize]{cleveref}
\crefname{section}{Sec.}{Secs.}
\Crefname{section}{Section}{Sections}
\Crefname{table}{Table}{Tables}
\crefname{table}{Tab.}{Tabs.}

\def\cvprPaperID{00004}
\def\confName{CVSport}
\def\confYear{2022}

\newcommand{\gva}[1]{#1\xspace}

\definecolor{cB}{RGB}{31,100,167}
\definecolor{cG}{RGB}{48,146,24}
\definecolor{cO}{RGB}{248,103,0}
\newcommand{\dB}[2]{\scriptsize{\color{cB}$#1\pm #2$}}

\newcommand{\dO}[2]{\scriptsize{\color{cO}$#1\pm #2$}}

\newcommand{\BallSeg}{\textit{BallSeg}\xspace}
\newcommand{\heatmap}{$\Psi$\xspace}
\newcommand{\mmdiam}{\rho}
\newcommand{\evaluation}{testing\xspace}
\newcommand{\model}{\mathcal{M}_{\theta}}

\begin{document}

\title{3D Ball Localization From A Single Calibrated Image}

\author{Gabriel Van Zandycke$^{1,2}$ \quad and \quad Christophe De Vleeschouwer$^1$\\[1mm]
\normalsize
$^1$ UCLouvain ICTEAM/ELEN Belgium \quad 
{\nolinkurl{<firstname>.<lastname>@uclouvain.be}}\\
\normalsize
$^2$ Sportradar \quad
{\nolinkurl{<f>.<lastname>@sportradar.com}}
}


\maketitle

\begin{abstract}
Ball 3D localization in team sports has various applications including automatic offside detection in soccer, or shot release localization in basketball.
Today, this task is either resolved by using expensive multi-views setups, or by restricting the analysis to ballistic trajectories.
\quad
In this work, we propose to address the task on a single image from a calibrated monocular camera by estimating ball diameter in pixels and use the knowledge of real ball diameter in meters.
This approach is suitable for any game situation where the ball is (even partly) visible.
To achieve this, we use a small neural network trained on image patches around candidates generated by a conventional ball detector. Besides predicting ball diameter, our network outputs the confidence of having a ball in the image patch.
\gva{Validations on 3 basketball datasets reveals that our model gives remarkable predictions on ball 3D localization. In addition, through its confidence output, our model improves the detection rate by filtering the candidates produced by the detector.}
\gva{The contributions of this work are (i) the first model to address 3D ball localization on a single image, (ii) an effective method for ball 3D annotation from single calibrated images, (iii) a high quality 3D ball evaluation dataset annotated from a single viewpoint.}
In addition, the code to reproduce this research will be made freely available at \quad {\url{https://github.com/gabriel-vanzandycke/deepsport}}

\end{abstract}


\section{Introduction}
\label{sec:intro}

Ball 3D localization has largely been studied~\cite{Kamble2019}, leading to industrial products like the automated line calling system in tennis or the goal line technology in soccer\cite{Bal2012}.
Most existing solutions rely on multiple view points~\cite{Kumar2011,Pingali2000,Ren2008,Lampert2012,Cheng2018,Parisot2011,Parisot2019,Maksai2016} to triangulate the 2D ball positions detected in individual frames, thereby offering high reliability in case of occlusions, which are frequent and severe in some team sports like basketball or American football.
Obviously, due to the exploitation of multiple synchronized cameras, those systems are expensive and complex to install, compared to single viewpoint setups.
Yet, there are many practical cases where ball 3D localization is valuable even though only a single viewpoint is available. This typically happens when the end-user is interested in
the 3D reconstruction and analysis of the play
e.g. to collect statistics related to the 3D movement of the ball and players.

Several works have addressed the question of 3D ball localization using single calibrated viewpoints. They generally rely on 3D ballistic trajectory fitting from the 2D detections~\cite{Chen2012,Shun2004,Chen2009,Kim1998,Skold2015,Parisot2019,Labayen2014,Silva2011}. This approach, very well described in~\cite{Skold2015}, is however limited to detections on video sequences (multiple consecutive images), when ball is in free fall (ballistic trajectories).

Nonetheless, in many game situations, the ball is only partially visible or not in free fall. Therefore, in our work, we drop assumptions of ballistic trajectory, time consistency, and clean visibility, and we aim at localizing the ball in 3D from a \emph{single} calibrated image.
\gva{In particular, we validate our method on basketball games where ballistic trajectories -- typically long passes and shots -- only represent less than 10\% of game time\footnote{Evaluated on our dataset presented in Section~\ref{sec:ourdataset}}.}
In short, our method estimates ball diameter in pixels and uses knowledge of the real ball size and camera calibration to locate the ball in the 3D space.
We use a two-stage approach where a small convolutional neural network (CNN) is trained to predict ball diameter from candidates generated by a detector. In addition, the CNN is also trained to confirm/refute the presence of a ball, thereby improving the performance of the global image-wise ball detector.
To the best of our knowledge, our approach is the first one to make ball 3D localization possible from a single viewpoint without resorting to temporal analysis and ballistic trajectory assumption. 

Since an appropriate dataset is crucial for the training of our model, we also present a method to produce high quality ball 3D annotations in single view setups, and used it to create a 3D localization dataset made freely available here: \url{https://www.kaggle.com/gabrielvanzandycke/ballistic-raw-sequences}.

\section{Method}
\label{sec:method}

We propose to formulate the ball 3D localization task in a single calibrated image as the combination of two image analysis problems, namely the ball detection and the ball diameter estimation problems.
Calibration information and knowledge of the actual ball diameter, in meters, are then used to compute the ball localization in the 3D space from its diameter and position in the image pixels space.

\subsection{Ball detection in image space}

To detect balls in the image space, we adopt the State-of-the-Art solution \BallSeg~\cite{VanZandycke2019} that uses a segmentation approach. Specifically, the model, based on an ICNet architecture~\cite{Zhao2018}, is trained to predict a mask of the ball. At test-time, the top-$k$ highest peaks are identified on the output heatmap \heatmap to produce $k$ ball candidates.

\subsection{Ball diameter estimation in pixels}
\label{sec:ourmethod}

\subparagraph{A baseline using a Hough circle transform.}
A natural and well-known approach to estimate the size of a circular and highly contrasted object in an image consists in applying the Hough-circle transformation (HCT)~\cite{Duda1972} to the image gradient magnitude. As a baseline, to estimate the ball diameter in the image space, we propose to use the Hough transformation on \heatmap, over a spacial neighborhood of the top ($k\textstyle{=}1$) ball candidate.

In practice, several pre-processing steps are applied prior to the Hough transformation. A morphological opening operation with a circular filter of diameter $\mmdiam$ is first applied to \heatmap. Edges are then detected using adaptive thresholding with hysteresis~\cite{Canny1986}, with high and low thresholds $\tau_h$ and $\tau_l$. Finally, the Hough-circle transformation is computed from candidate diameters bounded in a range defined by the physical limits of the scenes (see Section~\ref{sec:experiments}).

\subparagraph{Our proposed CNN-based method.}
In this work, we propose to adopt a Convolutional Neural Network (CNN) to regress the ball diameter, in pixels, from an image patch $I$ centered on the candidates proposed by any conventional ball detector.
For a detector that produces $k$ bounding boxes on the initial input image, when $k>1$, it is likely that most candidates are false positives. Hence, our CNN is trained to predict whether the input patch contains a ball or not, in addition to regressing the ball diameter.
As demonstrated in Section~\ref{sec:experiments}, this offers the possibility to increase the ball detection rate at constant false positive rate, compared to a detector configured to output the most confident detection.

The CNN model $\model$ adopted in this work follows a VGG~\cite{Simonyan2014} architecture that takes image patches $I$ as input, and extracts a feature vector which is connected to three fully connected layers inspired by Le~Cun \etal in~\cite{LeCun1998}. The last layer has 2 neurons that produce the regression $\hat{d}:=\model^d(I)$ and classification $\hat{c}:=\model^c(I)$ outputs.

\medskip

We trained $\model$ using batch stochastic gradient descent over $\theta$.
The classification output $\hat{c}$, is supervised with binary cross entropy loss
\begin{equation}
    \mathcal{L}^c(c,\hat{c}) = - c \cdot \log(\hat{c}) + (1-c) \cdot \log(1-\hat{c})\quad ,
\end{equation}
with $c\in \{0,1\}$ being the target class where $1$ means $I$ is centered on a ball.
The regression output is supervised with Huber~Loss~\cite{Huber1964} defined by
\begin{equation}
    \mathcal{L}^d(d,\hat{d}) = \begin{cases}
    \frac{1}{2}(d-\hat{d})^2 & \text{for}\quad |d-\hat{d}| \leq \delta \\
    \delta(|d-\hat{d}| - \frac{1}{2}\delta)& \text{otherwise}\quad ,
    \end{cases}
\end{equation}
where $d\in\mathbb{R}_0^{+}$ is the target diameter in pixels, and the meta-parameter $\delta$ represents the threshold at which the loss turns from linear to quadratic. The two losses were weighted with a parameter $\alpha$
\begin{equation}
    \mathcal{L} = \alpha\cdot\mathcal{L}^d + (1-\alpha)\cdot\mathcal{L}^c\quad ,
\end{equation} but $\mathcal{L}^d$ was ignored for non ball inputs.

\medskip

At test time, given $k$ candidates predicted by our detector, we extract $k$ crops $I_i$, with $i\in [1,k]$, from the initial RGB input image. The final ball detection candidate is the one that has maximum $c_i$.

\subsection{Turning ball diameter into 3D localization}

For a camera defined by a focal length $f$, pixel density $\mu_x$ and $\mu_y$ in $x$ and $y$ direction respectively, a skew coefficient $\gamma$, and a principal vector $(u_x, u_y)$, we have the camera matrix
\begin{equation}
    K := \begin{bmatrix}
    f\cdot\mu_x & \gamma & u_x \\
    0           & f\cdot\mu_y & u_y \\
    0           & 0          & 1
    \end{bmatrix}\quad .
\end{equation}
Given the diameter $d$ and ball position $(b_x,b_y)$ in pixels in the image space, the 3D rays of the ball center $\textbf{b}^c$ and two diametrically opposed ball edges $\textbf{e}_{-}^c$ and $\textbf{e}_{+}^c$, expressed in the camera coordinate system, are 
\begin{align}
    \textbf{b}^c &= K^{-1}\cdot\mathcal{R}\left( \begin{bmatrix}b_x\\b_y\\1\end{bmatrix}\right)\qquad\text{and}
    \\
    \textbf{e}_{\pm}^c &= K^{-1}\cdot\mathcal{R}\left( \begin{bmatrix}b_x\\b_y\textstyle{\pm}\frac{d}{2}\\1\end{bmatrix}\right)\quad ,
\end{align}
where $\mathcal{R}:\mathbb{R}^3\rightarrow\mathbb{R}^3$ is the function that rectifies image coordinates to handle lens distortions.

For a camera placed at $\textbf{c}^o\in\mathbb{R}^3$ in the world coordinates system, with an orientation defined by the rotation matrix $R$, and the true ball diameter $\phi$, the 3D localization in the ball is then given by $\textbf{b}^o$:
\begin{equation}
\label{eq:ball3d}
    \textbf{b}^o = R^T\cdot\frac{\phi\cdot \textbf{b}^c}{{\textbf{e}_{+}^c}_y - {\textbf{e}_{-}^c}_y}+\textbf{c}^o
\end{equation}

\section{Datasets}
\label{sec:dataset}

Our deep learning method requires a dataset to train and evaluate on. However, not many public datasets with 3D ball annotations exist. Most of the works mentioned in the introduction rely on private datasets, usually captured in one or few different environments~\cite{Cheng2018,Lampert2012,Maksai2016,Pingali2000,Ren2008,Silva2011,Skold2015}. To our knowledge, there are only three public datasets with annotated 3D ball position:

\textbf{Basket-APIDIS}~\cite{Vleeschouwer2008}
is a 15 minutes women basketball sequence captured by 7 un-synchronized cameras at about 22 frames per second in a single venue equipped with by a rather weak lighting setup, resulting in poorly contrasted images. 
Ball center was manually labelled by P. Parisot \etal on the 7 streams for 3 minutes of the game. A pseudo ground-truth of 3D localization was built by triangulation and time interpolation on a re-sampled stream of $800\times 600$ pixels at 25 frames per second~\cite{Parisot2019}. \gva{The difference with the acquisition frame rate implies alignment issues between the images and the corresponding ball 3D annotations due to duplicated frames.}

\textbf{Soccer-ISSIA}~\cite{DOrazio2009}
is a 2 minutes soccer sequence captured from 6 $1920\textstyle{\times}1080$ synchronized cameras, at 25 frames per second. Ball position was annotated with a semi automated process on the 6 streams and 3D localization can be retrieved using known calibration. However, as it is designed for player tracking, the ball is often out of the field of view when flying.

\begin{figure}[t]
    \centering
    \includegraphics[width=\columnwidth]{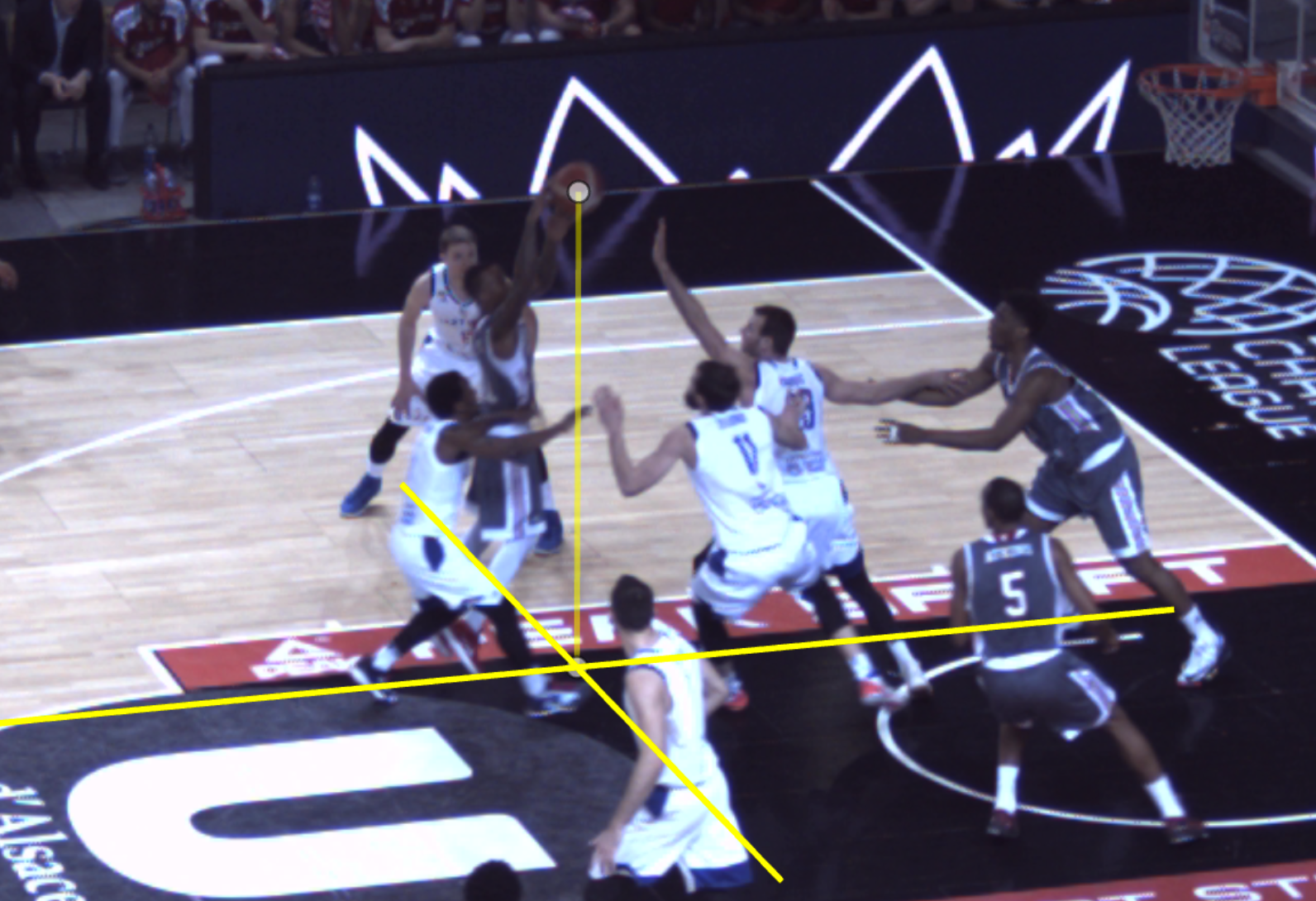}
    \caption{Annotation interface for ball 3D localization in calibrated images. A first point is clicked on the ball center, the second point is clicked at its vertical projection on the ground. The calibration is used to constrain the projection in a vertical line, and a cross aligned with the field geometry is drawn on the ground to help the annotator localizing the ball in the scene.}
    \label{fig:tool}
\end{figure}

\textbf{DeepSport}~\cite{DeepSportDataset, deepsport-dataset}
is a basketball dataset composed of 314 high-resolution panoramic images captured in 15 different basketball arenas during professional matches. Image resolution varies in each arena but generally reaches around $4500\textstyle{\times}1500$ pixels. \gva{Ball 3D annotation is given by the annotation of ball center and its vertical projection on the court in the image space (see Figure~\ref{fig:tool}).}
An extended version of the dataset containing 1200 additional annotated images of similar quality is available under non disclosure agreement (NDA) at~\cite{deepsport-dataset}.

\medskip

The soccer context does not offer the complexity encountered in basketball and the 2 minutes sequence from ISSIA dataset lacks in diversity. Hence, in this work, we focus on basketball, where balls are frequently occluded, often poorly contrasted, and suffer from noise and motion blur due to indoor lighting conditions. This makes it a challenging sport to address the ball 3D localization task.
Despite a scene captured by multiple viewpoints, the APIDIS dataset does not offer accurate ball 3D localization. Indeed, due to the lack of synchronization between the multiple image streams, we observed dozen of pixels differences between the 3D position projected in image space and the visual appearance of the ball. Additionally the dataset does not provide any information about ball visibility, in particular when it is hidden by the players as often encountered in basketball scenes.
Therefore, we decided to build on the DeepSport dataset for training, as it offers the largest environment diversity with different angles of view, ball color, court color, and lighting conditions.

\subsection{Annotation quality assessment}
\label{sec:ourdataset}
\gva{The lack of multiple viewpoints in the DeepSport dataset makes the 3D annotation quality uncertain. Yet, as our experiments presented in section~\ref{sec:loc3d} reveal, 
a few pixels diameter error can translate in meters in the 3D space.
Hence, we conducted a quality assessment of the annotations resulting from the approach adopted in~\cite{deepsport-dataset}, and compared it against a more natural approach that consists in annotating the ball diameter and using the relation defined by Eq.~\ref{eq:ball3d} to retrieve the 3D position.}




To build a ground-truth of 3D ball localization using a single calibrated camera, we propose to annotate multiple consecutive images when the ball follows a ballistic trajectory, and fit a motion model on the annotations to regularize and thereby denoise the initial 3D annotations.

\begin{figure}[h!]
    \centering
    \includegraphics[width=\columnwidth]{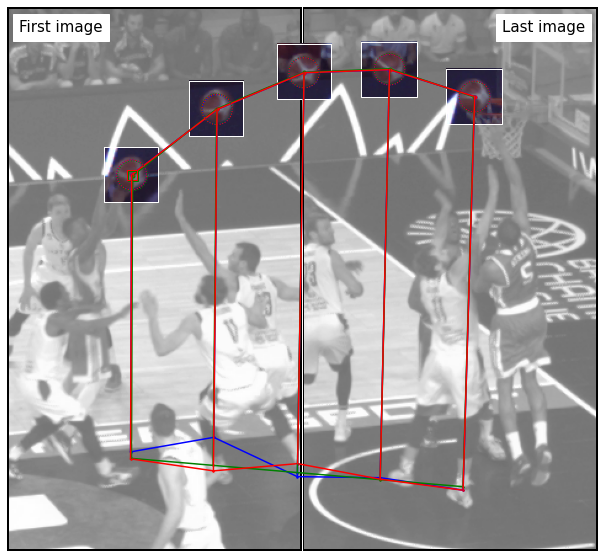}
    \caption{Process for achieving high quality ground-truth for ball 3D localization on calibrated images.
    (i) Ballistic trajectories are identified. 
    (ii) Images are annotated individually using the center+projection approach ({\color{blue}blue annotations}).
    (iii) Trajectories are visualized and inaccurate annotations are manually annotated again ({\color{red}red annotations}).
    (iv) A motion model is fitted to those annotations to obtain near perfection annotations ({\color{green!70!black}green annotations}).}
    \label{fig:trajectory}
\end{figure}

We captured 2 sequences during professional basketball games with a static calibrated monocular camera in 2 different arenas, and identified 35 ballistic trajectories composed of a total of 233 images. After an initial annotation using the center+projection approach, trajectories were individually inspected
\gva{on a visualization made from the first and last image of the trajectory, in which each annotation was displayed (see Figure~\ref{fig:trajectory}).
Annotations were then manually adjusted such that the trajectory derived from the motion model fitting would match the scene.}

\medskip

In practice, we used the second degree model $\mathcal{F}$, ignoring friction, ball spin and other effects:
\begin{equation}
    \mathcal{F}_{\textbf{p}_0, \dot{\textbf{p}}_0}(t) = \textbf{p}_0 + \dot{\textbf{p}}_0\ t - \begin{bmatrix}0\\0\\g\end{bmatrix} \frac{t^2}{2}\quad ,
\end{equation}
where $\textbf{p}_0$ and $\dot{\textbf{p}}_0$ are the initial position and initial velocity vectors, and $g=9.81 m/s^2$ is the gravity constant.

Given a trajectory composed of $N$ annotated positions $\hat{\textbf{p}}_i$ at times $t_i$, with $i\in[1,N]$, the motion model parameters are obtained by the following relationship
\begin{equation}
    \textbf{p}_0, \dot{\textbf{p}}_0 = \argmin_{\textbf{p}_0, \dot{\textbf{p}}_0} \sum_{i=1}^{N} \Big(\mathcal{F}_{\textbf{p}_0, \dot{\textbf{p}}_0}(t_i) - \hat{\textbf{p}}_i\Big)^2 \quad.
\end{equation}
Finally, the high quality positions $\textbf{p}_i$ are eventually computed by evaluating $\mathcal{F}_{\textbf{p}_0, \dot{\textbf{p}}_0}$ at $t=t_i$.

These high quality annotations can be considered as ground-truth to compare the distribution of errors resulting from the two annotation approaches. 
With a reasoning similar to Equation~\ref{eq:ball3d}, we can compute the diameter in pixels $d$ that corresponds to the position $\textbf{p}$, using the camera calibration and knowledge of the true diameter $\phi$.

\begin{figure}[h!]
    \centering
    \includegraphics[width=\columnwidth]{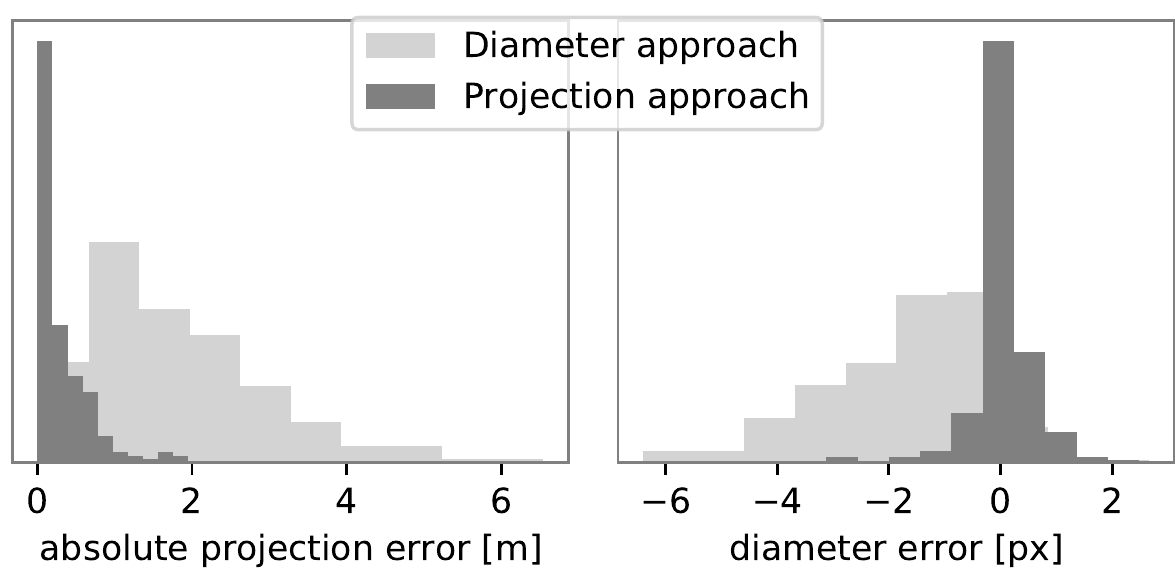}
    \caption{Comparison of two approaches to annotate ball 3D localization on single calibrated views. The diameter approach consists in annotating ball center and diameter, and the projection approach consists in annotating ball center and vertical projection on the ground. The diameter approach shows higher error and more specifically a tendency to under estimate the true diameter. The projection approach is more accurate and translates to a diameter error of less than 1 pixel.}
    \label{fig:ann}
\end{figure}

As shown in Figure~\ref{fig:ann}, the ball center+projection approach is significantly better for ball 3D annotation in single calibrated viewpoints.

\subsection{Used datasets}
As mentionned earlier, we used the DeepSport dataset for training, as it offers the largest environment diversity.
In contrast to the DeepSport dataset where many balls are partly occluded in cluttered scenes as often encountered in basketball games, our high quality evaluation set presented in Section~\ref{sec:ourdataset} is limited to samples where the ball moves freely, on ballistic trajectories and does not include occluded balls or balls held in player hands. For this reason, in the rest of the paper, the evaluation on the DeepSport dataset was generally preferred.
Nevertheless, the APIDIS dataset and our accurate evaluation set have been used to illustrate the generalization capabilities of our model and assess the impact of annotation errors. A summary of the different datasets used in this research is provided in Table~\ref{tab:datasets}, and samples of each dataset are given in Figure~\ref{fig:samples}.

\begin{table}[h]
    \small
    \centering
    \resizebox{\columnwidth}{!}{\begin{tabular}{lcccc}
    \toprule
         &  \!\! Images\!\!   & \!\! Scenes\!\! & \!\!\shortstack{ball size\\range {[px]}}\!\! & Resolution \\
\midrule
DeepSport~\cite{DeepSportDataset} &  314 & 15 & $14-37$ & $2-5$ Mpx \\
DeepSport ext.~\cite{deepsport-dataset} \kern-1em  & 1514 & 49 & $14-45$ & $2-5$ Mpx \\
Our evaluation set \kern-1em   &  233 &  2 & $19-40$ & \! $2336\times 1752$\! \\
APIDIS~\cite{Parisot2019}                    & 4019 &  1 &  $5-27$ & $800\times 600$ \\
    \bottomrule
    \end{tabular}
    }
    \caption{Summary of the different datasets used in this research. The DeepSport dataset was used both for training and for evaluation, as it offers the largest environment diversity with high quality annotations. The other datasets were used for evaluation only.
    The extended version of DeepSport was used to assess the impact of the training set size on the performances. 
    The numbers reported for the APIDIS dataset correspond to the images in which ball could be projected. Our evaluation set, presented in Section~\ref{sec:ourdataset}, is clean from any occlusions.}
\label{tab:datasets}
\end{table}

\pagebreak

\section{Experiments}
\label{sec:experiments}

Implementation details and metrics used to conduct this research are presented hereafter.

\subparagraph{Dataset}
As mentioned in Section~\ref{sec:dataset}, we use the DeepSport dataset for training, and have adopted the split defined in~\cite{DeepSportDataset}, where each fold contains images from basketball arenas exclusive to that fold. The fold \texttt{A} remained unseen and was used for \evaluation. In addition, we evaluated the models on the APIDIS dataset and our high quality evaluation set presented in Section~\ref{sec:dataset}.

\subparagraph{Ball detector} \BallSeg was trained as recommended in~\cite{VanZandycke2019}, with a random scaling strategy where ball size was kept between 14 and 37 pixels, corresponding to the range of ball sizes in the DeepSport dataset (see Table~\ref{tab:datasets}). 
To analyze the impact of working on smaller balls, we also considered halving the size range in specific trainings (see Section~\ref{sec:sizes}).
In our context, we are only interested in the ball being played. Therefore, only the top ($k=1$) candidate was kept, unless stated otherwise.

\subparagraph{Baseline method}
A $64\times 64$ pixels neighbourhood of \heatmap around the detection candidate was used to apply the different image processing steps; with $\mmdiam=37$, $\tau_l=10$ and $\tau_h=20$, selected by maximizing the performances on a validation set. The range of candidate diameters for the Hough transformation was constrained, with a 2 meters margin, by the physical boundary of the basketball court.

\subparagraph{Proposed method}
In practice, although our CNN method can be combined with any ball detector we used the detections produced by \BallSeg as input to our model, allowing a fair comparison with the baseline.
As described in Section~\ref{sec:ourmethod}, $\model$ is composed of a VGG feature extractor and 3 additional fully connected layers. We took VGG16 pre-trained on ImageNet~\cite{Deng2010} from TensorFlow library~\cite{tensorflow}, and used Glorot initializer~\cite{Glorot2010} to initialize the last layers.
Our CNN was trained using stochastic gradient descent over $\theta$, on batches of 4 different images, with $k=4$ candidates output by \BallSeg, making an effective batch size of 16 image patches $I$.
Training took 100 epochs using Adam optimizer~\cite{Kingma2015}, with a learning rate schedule starting at $10^{-4}$ and reducing by a factor 2 for 2 epochs every 10 epochs starting at epoch 50. The loss parameters $\delta=1.0$ and $\alpha=0.5$ were selected empirically using the validation set. Image patches of $64 \times 64$ pixels were used (see Section~\ref{sec:sl}).

\subparagraph{Metrics}
\label{sec:metrics}

To assess the diameter estimation accuracy, we used the mean absolute error (MAE$_\text{[px]}$) in pixels between the labels $d$ and the predictions $\hat{d}$. 
The corresponding projection error (MAE$_\text{[m]}$) is the distance in meters between the real position and the estimated position, computed with Equation~\ref{eq:ball3d}, projected on an horizontal plane. It depends on the camera pixel density, focal length and distance to the scene.
We also computed the relative distance error (MAE$_\text{[\%]}$) that represent the ratio between the projection error and the distance to the cameras.
To assess the ball detection performance, we used the Area under the Curve (AuC) associated to the Receiver Operating Characteristic (ROC) curve, plotting the True Positive (TP) rate against the False Positive (FP) rate for a varying detection threshold, as defined in~\cite{VanZandycke2019}. Here, the TP rate measures the fraction of images in which the ball is correctly detected while the FP rate measures the mean number of false candidates that are detected per image.

Experiments were repeated with $8$ different initializations of the networks (for both \BallSeg and the randomly initialized layers of our CNN). Unless stated otherwise, the mean metric over the 8 repetitions was reported, along with the associated standard deviation.

\subsection{Ball diameter estimation}

\begin{figure}[h]
    \centering
    \includegraphics[width=\columnwidth]{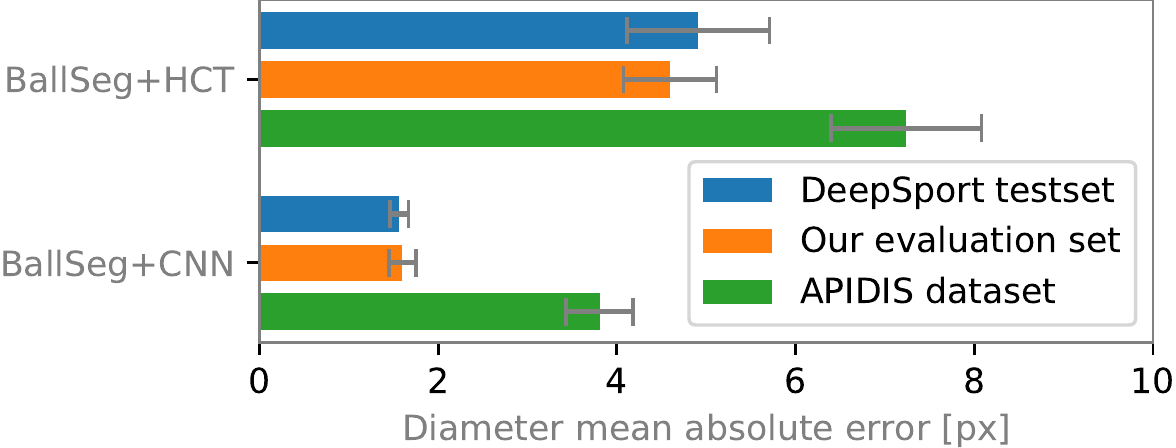}
    \caption{Our CNN method is compared against the Hough-circle transformation baseline (HCT) for diameter estimation on different Basketball evaluation sets. Our method reduces the mean absolute error by a factor two or more.}
    \label{fig:diameter}
\end{figure}

The comparison between the baseline using a Hough-circle transformation and our CNN approach is presented in Figure~\ref{fig:diameter}. 
In this setup, we evaluated on the true positive candidates produced by the detector \BallSeg.
We observe that our CNN-based approach reduces the mean absolute diameter estimation error MAE$_\text{[px]}$ by 3 pixels on each dataset, reaching less than 2 pixels on the DeepSport testset and our evaluation set.

The fact that the prediction accuracy does not improve between the DeepSport testset and our accurate evaluation set suggests that errors are not caused by annotation errors. Moreover, when referring to Figure~\ref{fig:ann}, we observe that a 2 pixels error is close to the estimated annotation error, meaning that our model achieves an accuracy comparable to human annotator. 


\subsection{Ball 3D localization error}
\label{sec:loc3d}
The estimated diameter in pixels, along with camera calibration and knowledge of the ball real diameter, allow computing the ball 3D position using Equation~\ref{eq:ball3d}. The error analysis given in Table~\ref{tab:table} shows that a diameter error (MAE$_\text{[px]}$) of $1.6$ pixels translates to a projection error (MAE$_\text{[m]}$) of less than two meters on our high-precision evaluation set. However this relationship highly depends on the camera pixel density, focal length and distance to the scene. Relative to the distance to the camera (MAE$_\text{[\%]}$), it corresponds to a 10\% error.
Recall that our image-based approach provides raw estimations, before any possible time filtering.



\begin{table}
    \footnotesize
    \centering
    \resizebox{\columnwidth}{!}{
    \begin{tabular}{lcccc}
    \toprule
         & TP$_\text{[\%]}$ & MAE$_\text{[px]}$ &  MAE$_\text{[m]}$ &  MAE$_\text{[\%]}$ \\
\midrule
         \multirow{2}{*}{\BallSeg + HCT}
& \dB{47}{7}  & \dB{4.9}{.8}&\dB{6.3}{1.0}&\dB{28}{5} \\
& \dO{83}{2}  & \dO{4.6}{.5}&\dO{5.1}{.5}&\dO{24}{4} \\
\midrule
         \multirow{2}{*}{\BallSeg + CNN}
& \dB{47}{7}  & \dB{1.6}{.1}&\dB{2.3}{.2}&\dB{10}{.9} \\
& \dO{83}{2}  & \dO{1.6}{.2}&\dO{1.8}{.2}&\dO{10}{.7} \\
\midrule
         \multirow{2}{*}{Oracle + CNN}
& \dB{100}{0} & \dB{1.9}{.1}&\dB{2.8}{.2}&\dB{12}{.6} \\
& \dO{100}{0} & \dO{1.5}{.1}&\dO{1.7}{.1}&\dO{10}{.5} \\
    \bottomrule
    \end{tabular}
    }
    \caption{Error analysis for three different approaches on two different evaluations sets: {\color{cB} DeepSport testset} first, and {\color{cO} Our~Evaluation~set} second.
    \BallSeg + HTC is the baseline that consists in applying Hough-circle transformation on the heatmap \heatmap output by \BallSeg. \BallSeg + CNN is our proposed method that estimate diameter or image patches $I$ around detection candidates generated by \BallSeg. Oracle + CNN is our proposed method applied on oracle detections instead of using a detector. The proportion of TP from the detector (TP$_\text{[\%]}$) is given for each dataset.
    The MAE$_\text{[px]}$ on oracle detections reaches 1.5 pixels on our evaluation set.
    }
\label{tab:table}
\end{table}

\subsection{Impact study of the detector}

As described in Section~\ref{sec:method}, we adopt a two-stage approach where the balls diameter estimated at the second stage depends on the detections proposed by the first stage.
To assess the performance of our CNN independently of the detection stage, 
the quality of diameter estimation was also evaluated on oracle detections by using input image patches $I$ around the true ball positions.
The error analysis given in Table~\ref{tab:table} shows that the performances using \BallSeg detections (1.6 pixels on both datasets) are close to the ones obtained using oracle detections (1.9 and 1.5 pixels).

This study reveals that our CNN gets the ability to correctly estimate the ball size on images where the ball is challenging to detect.
Moreover, the similarity between the DeepSport testset (subject to occlusions, see samples in~\cite{VanZandycke2019}) and on our evaluation set (free from occlusions), indirectly reveals that our method is robust to occlusions.

\subsection{Ball detection quality}

\begin{figure}[h]
    \centering
    \includegraphics[width=\columnwidth]{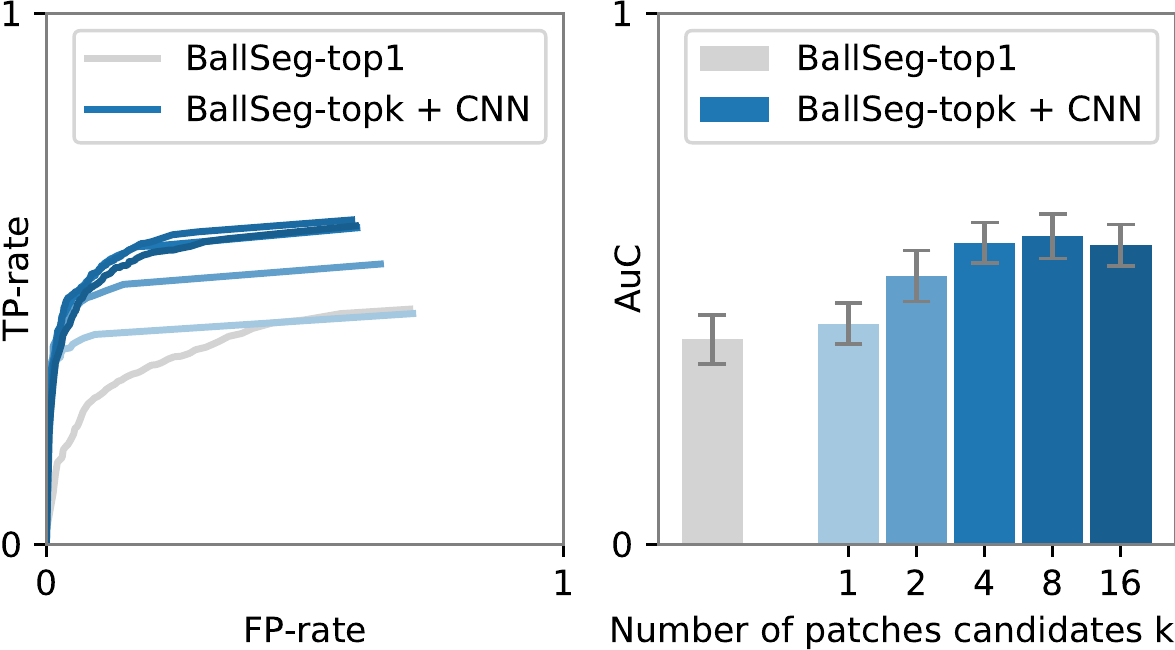}
    \caption{Impact of $k$ on the detection quality of \BallSeg-top$k$ combined with our CNN approach. (Left) ROC curves obtained by varying the detection threshold, on the accumulated detection candidates over the 8 repetitions (see Section~\ref{sec:metrics}). (Right) Area under the ROC curves for the different configurations. Our CNN is able to recover balls missed by \BallSeg-top1 but who ranked in the top$k$ with $k\geq 2$.}
    \label{fig:detection}
\end{figure}

With its classification output $\model^c$, our CNN model can be used to revise the detection confidence produced by the detector it is combined with. Furthermore, by considering $k$ ball candidates from the detector and keeping the candidate from image patch $I_j$ such that
\begin{equation}
    j = \argmax_{i\in[1,k]} \model^c(I_i)\qquad,
\end{equation}
our CNN becomes able to recover some of the detections missed when the detector adopts a top-1 detection strategy.
We conducted experiments with \BallSeg for different values of $k$ and, as revealed by Figure~\ref{fig:detection}, our model improves the detection quality by a large margin compared to \BallSeg-top1 alone.
More specifically, the ROC curves in Figure~\ref{fig:detection} reveal two positive effects.
First, our CNN is more effective in discriminating between true positives and false positives, as attested by the larger TP-rate at small FP-rate compared to \BallSeg for $k=1$.
Second, the number of true positives is largely increased with $k>1$.


\subsection{Impact of ball resolution.}
\label{sec:sizes}
\begin{figure}[t]
    \centering
    \includegraphics[width=\columnwidth]{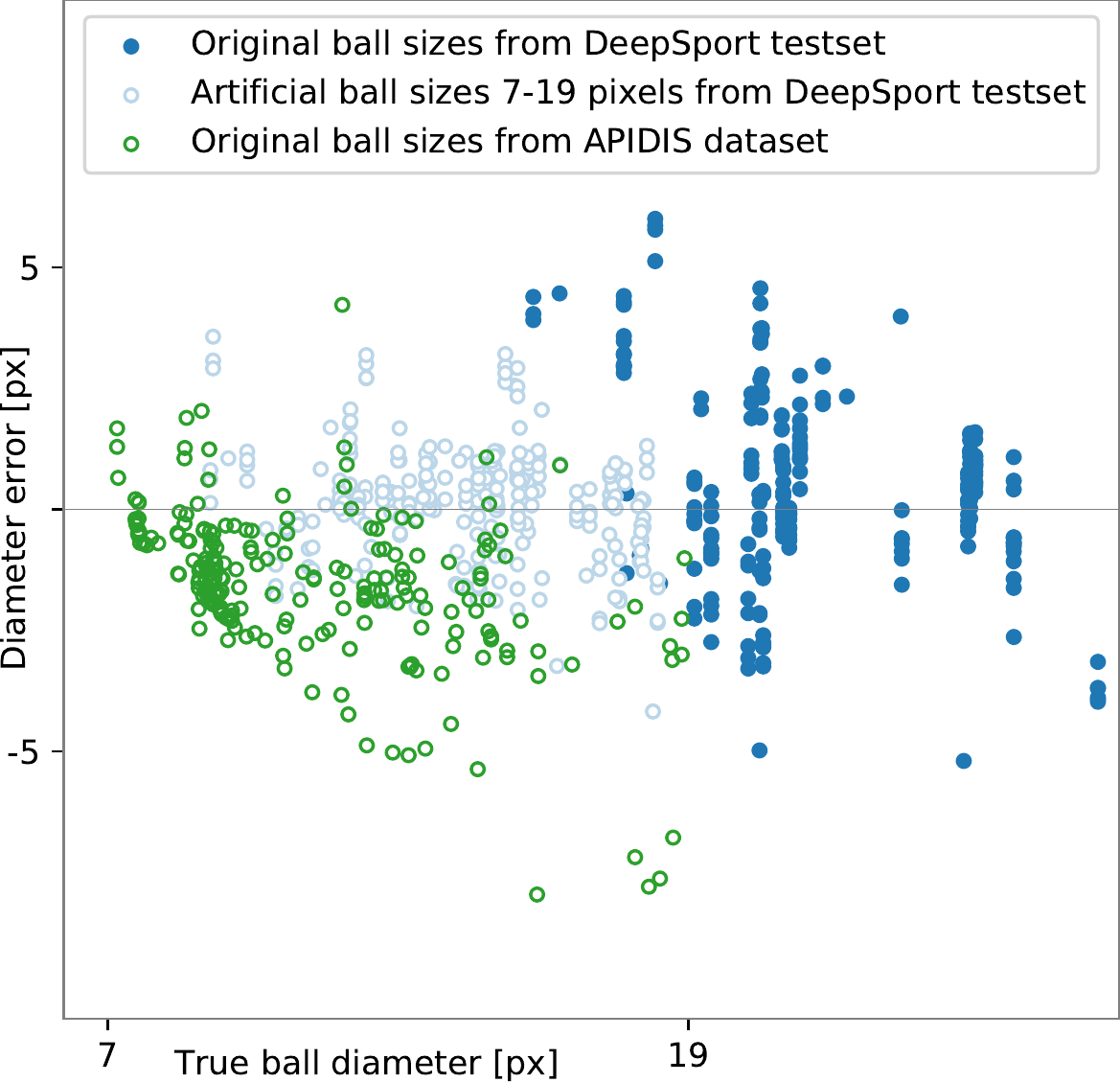}
    \caption{Diameter error distribution with respect to ball true diameter. Only 4 out of 8 repetitions (see Section~\ref{sec:metrics}) are displayed to declutter the figure. Our CNN was trained on half resolution images (empty circles) and evaluated both on the DeepSport testset where images were randomly scaled such that balls appear smaller, and the APIDIS dataset where balls larger than 19 pixels or smaller than 7 pixels were ignored. This experiments shows that our CNN approach performs very well at small diameters, validating the approach for many more applications. The distribution shift noticeable on the APIDIS dataset may be caused by the limited quality of that dataset (see Section~\ref{sec:dataset}).}
    \label{fig:scale2}
\end{figure}

In the DeepSport training dataset, ball sizes range from 14 to 37 pixels. However, they are many other applications for 3D ball localization where ball appears smaller, due to lower resolution images like in the APIDIS dataset, or simply because the physical size of the ball is smaller, as encountered for other sports like volleyball or tennis.
To evaluate the error distribution one could expect from smaller balls, we trained our CNN on half resolution images and evaluated it on sub-sampled images from the DeepSport testset where images were randomly scaled such that balls appear between 7 and 19 pixels.
Results presented in Figure~\ref{fig:scale2} reveal that the error distribution is centered around 0 and shows a standard deviation similar compared to the model working at normal resolution.
The model was also evaluated on the APIDIS dataset where balls are smaller (see Table~\ref{tab:datasets}).
Referring to Figure~\ref{fig:scale2}, it appears that the model underestimates the diameter from the balls of the APIDIS dataset.

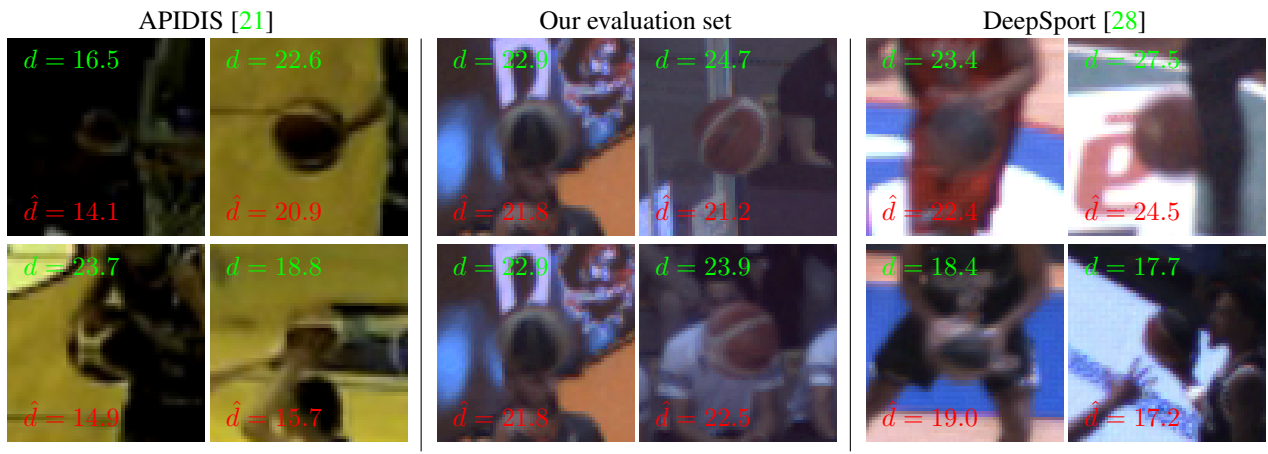
\begin{figure*}[t]
\centering
\begin{tabular}{c@{\;}c|c@{\;}c|c@{\;}c}
\multicolumn{2}{c}{APIDIS~\cite{Parisot2019}} &
\multicolumn{2}{c}{Our evaluation set} &
\multicolumn{2}{c}{DeepSport~\cite{DeepSportDataset}} \\
\begin{overpic}[width=.15\textwidth]{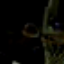}\put(8,85){\color{green}$d=16.5$}\put(8,8){\color{red}$\hat{d}=14.1$}\end{overpic}  &
\begin{overpic}[width=.15\textwidth]{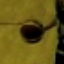}\put(8,85){\color{green}$d=22.6$}\put(8,8){\color{red}$\hat{d}=20.9$}\end{overpic}  &
\begin{overpic}[width=.15\textwidth]{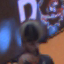}\put(8,85){\color{green}$d=22.9$}\put(8,8){\color{red}$\hat{d}=21.8$}\end{overpic}  &
\begin{overpic}[width=.15\textwidth]{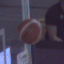}\put(8,85){\color{green}$d=24.7$}\put(8,8){\color{red}$\hat{d}=21.2$}\end{overpic}  &
\begin{overpic}[width=.15\textwidth]{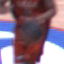}\put(8,85){\color{green}$d=23.4$}\put(8,8){\color{red}$\hat{d}=22.4$}\end{overpic}  &
\begin{overpic}[width=.15\textwidth]{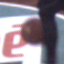}\put(8,85){\color{green}$d=27.5$}\put(8,8){\color{red}$\hat{d}=24.5$}\end{overpic}\\
\begin{overpic}[width=.15\textwidth]{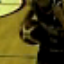}\put(8,85){\color{green}$d=23.7$}\put(8,8){\color{red}$\hat{d}=14.9$}\end{overpic}  &
\begin{overpic}[width=.15\textwidth]{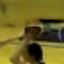}\put(8,85){\color{green}$d=18.8$}\put(8,8){\color{red}$\hat{d}=15.7$}\end{overpic}  &
\begin{overpic}[width=.15\textwidth]{assets/hqseq_0.png}\put(8,85){\color{green}$d=22.9$}\put(8,8){\color{red}$\hat{d}=21.8$}\end{overpic}  &
\begin{overpic}[width=.15\textwidth]{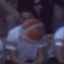}\put(8,85){\color{green}$d=23.9$}\put(8,8){\color{red}$\hat{d}=22.5$}\end{overpic}  &
\begin{overpic}[width=.15\textwidth]{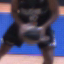}\put(8,85){\color{green}$d=18.4$}\put(8,8){\color{red}$\hat{d}=19.0$}\end{overpic}  &
\begin{overpic}[width=.15\textwidth]{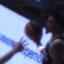}\put(8,85){\color{green}$d=17.7$}\put(8,8){\color{red}$\hat{d}=17.2$}\end{overpic}\\
\end{tabular}
    \caption{Image samples extracted from the three datasets. $d$ is the true ball diameter in pixels and $\hat{d}$ is the prediction from one of our models. Samples from the DeepSport dataset were extracted from the fold \texttt{A} (see Section~\ref{sec:experiments}). Samples from APIDIS dataset were scaled to match ball size range set during training.}
    \label{fig:samples}
\end{figure*}

Results presented in Figure~\ref{fig:scale2} reveal that our model handles smaller resolution balls from the DeepSport dataset very well, validating the approach for many more applications.
We conjecture that the distribution shift observed on the APIDIS dataset is caused by the poorly contrasted images of that dataset (see Figure~\ref{fig:samples}).

\subsection{Impact of image patch size}
\label{sec:sl}
\begin{figure}[h]
    \centering
    \includegraphics[width=\columnwidth]{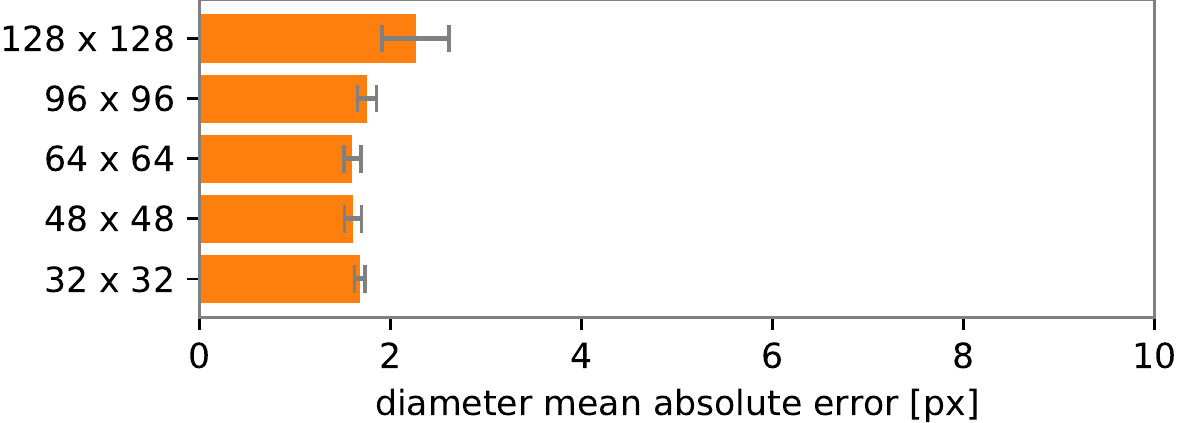}
    \caption{Diameter estimation evaluated on different input image patch $I$ sizes. This study was conducted on oracle detections from our high-accuracy evaluation set, clean from any occlusions.}
    \label{fig:sl}
\end{figure}

Multiple patch sizes were considered. As the Figure~\ref{fig:sl} reveals, $64\times 64$ image patches resulted in the most accurate estimation of the ball diameter. This study was conducted on oracle detections from our high-accuracy evaluation set, clean from any occlusions.

\subsection{Impact of training set size}
\begin{figure}[h]
    \centering
    \includegraphics[width=\columnwidth]{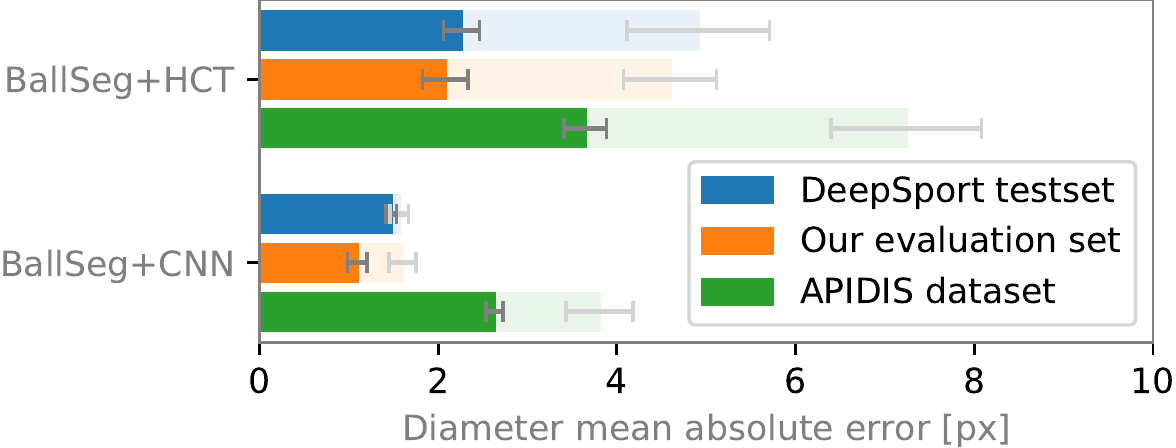}
    \caption{Evaluation of diameter estimation when training the models (both \BallSeg and our CNN) on the extended DeepSport dataset (see Table~\ref{tab:datasets}). Results from Figure~\ref{fig:diameter} are shown dimmed for an easier comparison. 
    BallSeg+HCT sees a huge performance gain from being trained on a larger dataset, but our CNN model performs better with a pixel error down to one pixel on our evaluation set.}
    \label{fig:244fr}
\end{figure}

We also evaluated our method by training both the detector \BallSeg and our CNN on the extended DeepSport dataset shared under NDA by~\cite{deepsport-dataset} (see Table~\ref{tab:datasets}). 
Figure~\ref{fig:244fr} reveals that our CNN benefits from being trained on a larger dataset, but also that the baseline experiences a significant gain with a drop of more than 50\% of the mean absolute diameter error on all datasets. Visual inspection of the output heatmap \heatmap shows that ball is much better segmented, resulting in a better diameter estimation using the Hough circle transform approach.
Evaluation on the APIDIS dataset consistently yields weaker performances, corroborating the observation that this dataset suffers from low contrasted images and poor annotations.

\section{Discussion}

The method presented in this work achieves a mean absolute diameter error of 1.6 pixels in the best conditions (ball free from occlusion on high resolution images), which translates to a projection error of 1.8 meters. 
This might appear as being a lousy approximation. However, in many applications (like 3D analysis of the play to collect statistics), having an automated estimation of ball 3D localization from a single camera can be extremely valuable, even with an imprecision of two meters.
Besides, our image-based approach provides raw estimations that could be improved based on temporal filtering when needed. 
However, investigating such temporal regularization goes beyond the scope of our paper.
Additionally, for industrial applications, collecting a dedicated dataset is generally required. And, as our experiments reveal, there is margin for improvements by training on a larger dataset.

\section{Conclusion}
\label{sec:concl}

This work has presented the first method to localize a ball in the 3D space from a single monocular image. Our approach is not restricted to balls following a ballistic trajectory, and works on any image where the ball is (event partly) visible, provided that we have camera calibration (which is mandatory for 3D localization) and know the real ball diameter (which is generally the case for many applications).
Our method combines a small CNN with an arbitrary ball detector, and compares favorably against a baseline using classical image processing techniques. Additionally, our CNN improves the detection capabilities of the detector it is combined with.

To conduct our study, we have also compared two methods to annotate the ball 3D position on a single image. The results of this comparison demonstrated the benefit of annotating the ball center and its projection on the ground compared to a natural approach of annotating ball diameter.

\section*{Acknowledgement}
Part of this work has been funded by the Walloon Region project DeepSport and by Synergy Sports Technology, a division of Sportradar.
C. De Vleeschouwer is a Research Director of the Fonds de la Recherche Scientifique - FNRS.

{\small
\bibliographystyle{ieee_fullname}
\bibliography{ball3d}

\begin{thebibliography}{10}\itemsep=-1pt

\bibitem{Bal2012}
Baljinder Bal and Gaurav Dureja.
\newblock {Hawk Eye: A Logical Innovative Technology Use in Sports for
  Effective Decision Making}.
\newblock {\em Sport Science Review}, 21(1-2):107--119, may 2012.

\bibitem{Canny1986}
John Canny.
\newblock {A Computational Approach to Edge Detection}.
\newblock {\em IEEE Transactions on Pattern Analysis and Machine Intelligence},
  PAMI-8(6):679--698, 1986.

\bibitem{Chen2009}
Hua~Tsung Chen, Ming~Chun Tien, Yi~Wen Chen, Wen~Jiin Tsai, and Suh~Yin Lee.
\newblock {Physics-based ball tracking and 3D trajectory reconstruction with
  applications to shooting location estimation in basketball video}.
\newblock {\em Journal of Visual Communication and Image Representation},
  20(3):204--216, 2009.

\bibitem{Chen2012}
Hua~Tsung Chen, Wen~Jiin Tsai, Suh~Yin Lee, and Jen~Yu Yu.
\newblock {Ball tracking and 3D trajectory approximation with applications to
  tactics analysis from single-camera volleyball sequences}.
\newblock {\em Multimedia Tools and Applications}, 60(3):641--667, 2012.

\bibitem{Cheng2018}
Xina Cheng, Norikazu Ikoma, Masaaki Honda, and Takeshi Ikenaga.
\newblock {Simultaneous physical and conceptual ball state estimation in
  volleyball game analysis}.
\newblock {\em 2017 IEEE Visual Communications and Image Processing, VCIP
  2017}, 2018-Janua:1--4, 2018.

\bibitem{Deng2010}
Jia Deng, Wei Dong, Richard Socher, Li-Jia Li, {Kai Li}, and {Li Fei-Fei}.
\newblock {ImageNet: A large-scale hierarchical image database}.
\newblock pages 248--255. Institute of Electrical and Electronics Engineers
  (IEEE), mar 2010.

\bibitem{tensorflow}
TensorFlow Developers.
\newblock {TensorFlow}.
\newblock feb 2022.

\bibitem{DOrazio2009}
T. D'Orazio, M. Leo, N. Mosca, P. Spagnolo, and P.~L. Mazzeo.
\newblock {A semi-automatic system for ground truth generation of soccer video
  sequences}.
\newblock In {\em 6th IEEE International Conference on Advanced Video and
  Signal Based Surveillance, AVSS 2009}, pages 559--564, 2009.

\bibitem{Duda1972}
Richard~O. Duda and Peter~E. Hart.
\newblock {Use of the Hough Transformation to Detect Lines and Curves in
  Pictures}.
\newblock {\em Communications of the ACM}, 15(1):11--15, jan 1972.

\bibitem{Glorot2010}
Xavier Glorot and Yoshua Bengio.
\newblock {Understanding the difficulty of training deep feedforward neural
  networks}.
\newblock In {\em Journal of Machine Learning Research}, volume~9, pages
  249--256, 2010.

\bibitem{Huber1964}
Peter~J. Huber.
\newblock {Robust Estimation of a Location Parameter}.
\newblock {\em The Annals of Mathematical Statistics}, 35(1):73--101, mar 1964.

\bibitem{Kamble2019}
Paresh~R. Kamble, Avinash~G. Keskar, and Kishor~M. Bhurchandi.
\newblock {Ball tracking in sports: a survey}.
\newblock {\em Artificial Intelligence Review}, 52(3):1655--1705, 2019.

\bibitem{Kim1998}
Taeone Kim, Yongduek Seo, and Ki~Sang Hong.
\newblock {Physics-based 3D position analysis of a soccer ball from monocular
  image sequences}.
\newblock {\em Proceedings of the IEEE International Conference on Computer
  Vision}, pages 721--726, 1998.

\bibitem{Kingma2015}
Diederik~P. Kingma and Jimmy~Lei Ba.
\newblock {Adam: A method for stochastic optimization}.
\newblock In {\em 3rd International Conference on Learning Representations,
  ICLR 2015 - Conference Track Proceedings}. International Conference on
  Learning Representations, ICLR, dec 2015.

\bibitem{Kumar2011}
K.~C.Amit Kumar, Pascaline Parisot, and Christophe {De Vleeschouwer}.
\newblock {Demo: Spatio-temporal template matching for ball detection}.
\newblock In {\em 2011 5th ACM/IEEE International Conference on Distributed
  Smart Cameras, ICDSC 2011}, 2011.

\bibitem{Labayen2014}
Mikel Labayen, Igor~G. Olaizola, Naiara Aginako, and Julian Florez.
\newblock {Accurate ball trajectory tracking and 3D visualization for
  computer-assisted sports broadcast}.
\newblock {\em Multimedia Tools and Applications}, 73(3):1819--1842, dec 2014.

\bibitem{Lampert2012}
Christoph~H. Lampert and Jan Peters.
\newblock {Real-time detection of colored objects in multiple camera streams
  with off-the-shelf hardware components}.
\newblock {\em Journal of Real-Time Image Processing}, 7(1):31--41, 2012.

\bibitem{LeCun1998}
Yann LeCun, L{\'{e}}on Bottou, Yoshua Bengio, and Patrick Haffner.
\newblock {Gradient-based learning applied to document recognition}.
\newblock {\em Proceedings of the IEEE}, 86(11):2278--2323, 1998.

\bibitem{Maksai2016}
Andrii Maksai, Xinchao Wang, and Pascal Fua.
\newblock {What players do with the ball: A physically constrained interaction
  modeling}.
\newblock In {\em Proceedings of the IEEE Computer Society Conference on
  Computer Vision and Pattern Recognition}, volume 2016-Decem, pages 972--981,
  2016.

\bibitem{Parisot2011}
Pascaline Parisot and Christophe {De Vleeschouwer}.
\newblock {Graph-based filtering of ballistic trajectory}.
\newblock In {\em Proceedings - IEEE International Conference on Multimedia and
  Expo}, 2011.

\bibitem{Parisot2019}
Pascaline Parisot and Christophe {De Vleeschouwer}.
\newblock {Consensus-based trajectory estimation for ball detection in
  calibrated cameras systems}.
\newblock {\em Journal of Real-Time Image Processing}, 16(5):1335--1350, 2019.

\bibitem{Pingali2000}
Gopal Pingali, Agata Opalach, and Yves Jean.
\newblock {Ball tracking and virtual replays for innovative tennis broadcasts}.
\newblock {\em Proceedings-International Conference on Pattern Recognition},
  15(4):152--156, 2000.

\bibitem{Ren2008}
Jinchang Ren, James Orwell, Graeme~A. Jones, and Ming Xu.
\newblock {Real-time modeling of 3-D soccer ball trajectories from multiple
  fixed cameras}.
\newblock {\em IEEE Transactions on Circuits and Systems for Video Technology},
  18(3):350--362, 2008.

\bibitem{Shun2004}
Hubert Shum and Taku Komura.
\newblock {A spatiotemporal approach to extract the 3D trajectory of the
  baseball from a single view video sequence}.
\newblock Technical report, 2004.

\bibitem{Silva2011}
Hugo Silva, Andr{\'{e}} Dias, Jos{\'{e}} Almeida, Alfredo Martins, and Eduardo
  Silva.
\newblock {Real-Time 3D Ball Trajectory Estimation for RoboCup Middle Size
  League Using a Single Camera}.
\newblock {\em Lecture Notes in Computer Science (including subseries Lecture
  Notes in Artificial Intelligence and Lecture Notes in Bioinformatics)}, 7416
  LNCS:586--597, 2011.

\bibitem{Simonyan2014}
Karen Simonyan and Andrew Zisserman.
\newblock {Very deep convolutional networks for large-scale image recognition}.
\newblock In {\em 3rd International Conference on Learning Representations,
  ICLR 2015 - Conference Track Proceedings}. International Conference on
  Learning Representations, ICLR, sep 2015.

\bibitem{Skold2015}
Jonas Sk{\"{o}}ld.
\newblock {Estimating 3D-trajectories from Monocular Video Sequences Estimating
  3D-trajectories from Monocular Video Sequences}.
\newblock 2015.

\bibitem{DeepSportDataset}
Gabriel {Van Zandycke}.
\newblock {DeepSport dataset:
  \url{https://www.kaggle.com/gabrielvanzandycke/deepsport-dataset}}, 2021.

\bibitem{VanZandycke2019}
Gabriel {Van Zandycke} and Christophe {De Vleeschouwer}.
\newblock {Real-time CNN-based segmentation architecture for ball detection in
  a single view setup}.
\newblock {\em MMSports 2019 - Proceedings of the 2nd International Workshop on
  Multimedia Content Analysis in Sports, co-located with MM 2019}, pages
  51--58, 2019.

\bibitem{Vleeschouwer2008}
C~De Vleeschouwer, Fan Chen, and Damien Delannay.
\newblock {Distributed video acquisition and annotation for sport-event
  summarization}.
\newblock {\em NEM summit}, 2008.

\bibitem{deepsport-dataset}
Gabriel~Van Zandycke.
\newblock {Image and Signal Processing Group (UCL) | Softwares:
  \url{https://sites.uclouvain.be/ispgroup/Softwares/DeepSport}}.

\bibitem{Zhao2018}
Hengshuang Zhao, Xiaojuan Qi, Xiaoyong Shen, Jianping Shi, and Jiaya Jia.
\newblock {ICNet for Real-Time Semantic Segmentation on High-Resolution
  Images}.
\newblock In {\em Lecture Notes in Computer Science (including subseries
  Lecture Notes in Artificial Intelligence and Lecture Notes in
  Bioinformatics)}, volume 11207 LNCS, pages 418--434. Springer Verlag, 2018.

\end{thebibliography}
}

\end{document}